\documentclass[10pt, a4paper]{article}

\usepackage[final]{lrec2026} 
\usepackage{booktabs}
\usepackage{multirow}
\usepackage{enumitem}
\usepackage{colortbl}
\usepackage{xcolor}
\usepackage{tabularx}

\title{Whose Voice Counts? \\ Mapping Stakeholder Perspectives on AI Through Public Submissions to the U.S. Government}

\name{Alina Karakanta\textsuperscript{1}, Alex Christiansen\textsuperscript{2}, Tomás Dodds\textsuperscript{3}, Bissie Anderson\textsuperscript{2},\\ { \large\textbf{Matteo Fuoli\textsuperscript{2}}, \textbf{Marcus Perlman\textsuperscript{2}}, \textbf{Aletta G. Dorst\textsuperscript{1}}}}
\address{
  \textsuperscript{1}Leiden University Centre for Linguistics, Leiden University \\
  \textsuperscript{2} Department of Linguistics and Communication, University of Birmingham\\ 
  \textsuperscript{3}School of Journalism and Mass Communication, University of Wisconsin-Madison\\
\texttt{a.karakanta@leidenuniv.nl}
}

\abstract{
As artificial intelligence (AI) systems become more common in our daily lives, it is important to understand how different stakeholders comprehend and envisage the role that these technologies play in shaping social, political, and economic realities. In this paper, we investigate public perceptions of AI based on a corpus of letters submitted during the public consultation for the Trump Administration's US AI Action Plan. To this aim, we release a corpus cleaning pipeline and perform topic modelling and frequency analysis to explore predominant topics discussed by different subgroups (e.g., academia, individuals, private sector) and those appearing in the AI Action Plan. Our results show that individuals voice strong concerns related to the impact of AI on life, while other stakeholders are more concerned with AI development. Our comparison of topics suggests that the AI Action Plan reflects predominantly the concerns of the private sector on security, policies, and development, with individuals' concerns less represented.
 \\ \newline  }

\begin{document}

\maketitleabstract

\section{Introduction}

In January 2025, Donald Trump signed an executive order aimed at removing “barriers to American leadership in artificial intelligence.” The goal, according to the same order, was to “sustain and enhance America’s global AI dominance in order to promote human flourishing, economic competitiveness, and national security” \cite{whitehouse2025removing}
. In a subsequent Request for Information
, issued in February 2025, the United States Office of Science and Technology Policy (OSTP) requested input from all interested parties on the Development of an Artificial Intelligence Action Plan. The Action Plan aimed to “define the priority policy actions needed to sustain and enhance America’s AI dominance, and to ensure that unnecessarily burdensome requirements do not hamper private sector AI innovation” \cite{nitrd2025comments}. As part of the consultation, the OSTP collected input in the form of public letters from different stakeholders, including individual actors, academia, industry, non-federal state actors, non-profit organisations and the private sector.

While the current administration of the US government has been quick to announce its interest in integrating AI technologies into healthcare, education, science, law and the military, it is yet an open question whether a wider set of stakeholders 
shares this desire for AI integration and domination, and whether the beliefs, concerns, and suggestions expressed during the consultation are ultimately reflected in the AI action plan. 

Drawing on 10,068 letters sent to the OSTP in response to their request, this paper examines stakeholders’ sentiment towards AI to create a comprehensive view of how different actors perceive the social, political and economic role of AI in today’s society. We first build and publicly release a processing pipeline to compile a clean corpus from the letters and create subcorpora based on the six categories of stakeholders that responded to the government’s request (academia, individuals, associations, non-federal, non-profit, private sector), as well as the 23-page AI Action Plan. Then, we conduct topic modelling to identify the topics and themes that are more prevalent for each group. Lastly, we compare the topics and their representation with frequent words in the AI Action Plan, to identify which topics and by which stakeholders are reflected in the Executive Order. We ask the following research questions:  

\textbf{RQ1}: What are the prominent topics discussed by each of the six stakeholder groups?

\textbf{RQ2}: To what extent are the suggestions and concerns put forth by each of the six stakeholder groups reflected in the AI Action Plan?

Understanding how different stakeholders perceive and communicate the role of AI to the government is important, as these concerns can later translate into defining regulatory provisions with potentially severe repercussions \cite{dodds2025old}. Moreover, governments strategically leverage public perception and lobbying efforts to build consensus and legitimacy for policy decisions, underscoring the need to examine whose voices are counted in the policymaking process, and who is disregarded.

In this unprecedented time of change brought about by AI, this study provides new evidence on public perceptions of AI, its democratic futures, and possible regulatory approaches. Our study is uniquely positioned at this point in AI’s evolution. Here, we capture how different parts of American society feel about these systems in a period that many observers have characterised as an AI investment bubble, comparable to other revolutionary innovations that have also gone through hype cycles, yet with far more immediate and widespread societal impact. Our research addresses this urgency by examining authentic policy communications during a pivotal consultation process, offering insights into whose concerns are amplified or marginalised in shaping America's AI governance at a moment when such decisions will have profound implications for decades to come. 

\section{Related work}
\label{sec:related}

\subsection{Perceptions on artificial intelligence}
Understanding public opinions on AI has been a subject of growing interest in recent years, with multiple surveys conducted both at focused and national scales. 
Most studies found a general ambivalence of US respondents towards AI. In national US surveys, Americans expressed mixed support for the development of AI 
\cite{zhang2019artificial} and the view that AI will be either good or bad for society, depending on what happens \cite{vu2022effects}. While most respondents believed that the development of AI should be encouraged 
as AI will simplify life and increase efficiency, around 20\% of respondents believed that AI may destroy humanity eventually \cite{bozkurt2025artificial}.
Despite this ambivalence, all surveys point towards similar patterns of concerns, such as the violation of privacy and civil liberties, disinformation, cybersecurity and data privacy \cite{zhang2019artificial}, as well as increases in unemployment and social inequalities \cite{rainie2022americans,dupont2023does,bozkurt2025artificial}.

Another point of agreement among Americans 
was that AI should be carefully managed \cite{zhang2019artificial}. However, the trust in the institutions implementing AI varies depending on the type of institution. 
Americans recognised that governance is important for developing AI but had low to moderate levels of trust in governmental, corporate, and multistakeholder institutions to manage AI in the public’s interest. Public trust varied greatly, with Americans trusting university researchers and the US military more, followed by tech companies (excluding Facebook), US federal or state governments, and Facebook coming last \cite{zhang2020us}. 


Most of these studies were conducted in the pre-ChatGPT era (launched in November of 2022), when the impact AI could have on our daily lives was seen as something for the distant future. Since the widespread adoption of large language models and generative AI, public discourse has shifted dramatically, with AI transitioning from a theoretical concern to an immediate reality affecting employment, education, creative industries, and governance.

\subsection{Corpus-based studies}
While surveys provide a direct way to investigate perceptions on AI, corpora are another rich source of views on AI. The few corpus-based studies available have mostly focused on how news articles and journalists perceive the hype around artificial intelligence.
In a study investigating news data between 1958-2018, \citet{garvey2020sentiment} found that sentiment in news on AI was not negative until 2018. \citet{fast2017long}'s study on a corpus of articles on AI from the New York Times over a 30-year period found an increase in mentions of AI after 2009. More positive perceptions were detected but growing negative perceptions were found on specific topics, such as loss of control of AI, ethical concerns, and the negative impact of AI on work. 
Benefits of AI were discussed more frequently than its risks \cite{chuan2019framing}, who analysed frames a small number of articles from more newspapers. Business and technology were the primary topics discussed around AI in news. More optimistic than pessimistic views on AI were found in \cite{cools2024exactly}, who investigated \textit{Automation} and \textit{AI} in two US newspapers 
over 35 years through topic modelling and manual framing. AI was mainly discussed through the topics of \textit{Work}, \textit{Art}, and \textit{Education}. Both studies mention that ethics or moral issues were discussed as main themes, even though \citet{ouchchy2020ai} found that media coverage on AI ethics was still shallow.


These works offered a quite uniform view of how AI is framed in news; however, a 12-country study which analysed the news coverage of AI in the Global North and South found significant differences in the framing of AI \cite{ittefaq2025global}. Global North newspapers gave less coverage to AI solutions, while AI regulations received limited coverage from the Global South newspapers. 
Global South media stressed AI-driven global/regional economic cooperation, while the Global North focused on countering China/Russia threats, potentially fuelling the rhetoric on race for strategic advantage on AI \cite{cave2018ai}.

The domain closer to our study is perspectives based on Twitter data \cite{miyazaki2024public}, however this study focused on AI perceptions based on occupation and usage. No study has explored public opinions through letters to the government. 

\section{Methodology}
\subsection{Corpus processing}
The data for this analysis comes from the Networking and Information Technology Research and Development (NITRD) website.\footnote{https://www.nitrd.gov/coordination-areas/ai/90-fr-9088-responses/} In total, 10,068 responses were submitted to the RFI, written in English, and classified by NITRD in the following respondent groups: academia, individuals, industry/professional/scientific associations, non-federal government (including state, local, and tribal governments), non-profit organisations, and private sector organizations. The website allows for filtering by respondent type and searching for respondents, but does not provide any options for exploring topics or tendencies inside these groups. Thus, we created a pipeline to clean and process the data. The code to obtain the clean version of the corpus is publicly available at \url{https://github.com/fatalinha/AI-Action-Plan}. Below, we outline the data processing steps:
\begin{itemize}[leftmargin=5pt]
    \item \textbf{File download}: The individual responses were downloaded as .zip file from the NITRD website. This contains all individual responses as separate pdf files.
    \item \textbf{Text extraction}: The pdf files were converted into txt using pdftotext\footnote{https://www.cyberciti.biz/faq/converter-pdf-files-to-text-format-command/}, a library which is part of poppler-utils. Except for the input/output filenames, no other options were passed.
    \item \textbf{Sentence splitting}: The obtained .txt files contained new line symbols at the end of each line. Therefore, the lines needed to be merged to full sentences. For each paragraph in the .txt files, identified by two consecutive new lines, the text was first merged and then sentence-split using the Stanza toolkit, version 1.10.1 \cite{qi2020stanza}.
    \item \textbf{Data cleaning}: The data required cleaning to remove traces of the pdf-to-text and email boilerplate. First, encoding issues were fixed by converting the text to utf-8. Quotes and other special characters were normalised. Any corrupted files that could not be recovered were deleted. Email boilerplate such as sender details, headers, footers and confidentiality notices was removed by identifying recurring sentences in the email template. Lastly, emails, urls, and footnotes were removed using regular expressions.
    \item \textbf{Division by respondent type}: Subcorpora were created based on respondent type.
\end{itemize}

In addition to the responses, our pipeline includes a clean version of the AI Action Plan.\footnote{https://www.whitehouse.gov/wp-content/uploads/2025/07/Americas-AI-Action-Plan.pdf} To ensure consistency, we applied the same cleaning process: pdf-to-text conversion, sentence splitting and cleaning. This allows for comparing the discourse and topics present in the responses and the final Act in a coherent way. 

The statistics of the resulting corpus are shown in Table~\ref{tab:corpus-stats}. Responses from individuals correspond to more than 90\% of the collected responses and 50\% of the words in the corpus. Among the rest of the stakeholders, the largest subcorpora come from the private sector, followed by non-profit organisations, industrial professional scientific organisations, and academia and non-federal organisations.

\begin{table}[ht]
\begin{small}
    \centering
    \begin{tabular}{l|r|r|r}\toprule
         & Docs & Sents & Tokens \\  \midrule
    Academia     & 81 & 10,626 & 235,065\\
    Individuals & 9,308 & 91,994 & 1,948,013\\
    Indus/Prof/Scient. & 178 & 18,806 & 424,832\\
    Non-federal & 10 & 492 & 12,911\\
    Non-profit & 192 & 25,401 & 535,668\\
    Private sector & 292 & 31,541 & 671,216\\ \midrule
    AI Action Plan & 1 & 232 & 10,825 \\ \bottomrule
    
    \end{tabular}
    \caption{Statistics of the corpus per respondent type, as well as the AI Action Plan.}
    \label{tab:corpus-stats}
    \end{small}
\end{table}

\begin{table*}[!t]
\begin{small}
    \centering
    \begin{tabular}{@{}*2l@{}} \toprule
     \textbf{Interpretation} & \textbf{Words} \\ \midrule
    \rowcolor{black!5}Energy & generative,	water,	ai,	texas,	artists,	power,	energy,	data,	art,	work\\ 
    Stop theft & stop,	people,	don,	steal,	theft,	copyright,	work,	just,	artists,	creative\\ 
    \rowcolor{black!5}Openai & openai,	ai,	copyright,	open,	theft,	steal,	people,	companies,	use,	models\\ 
    Future & future,	believe,	holds,	place,	profits,	ai,	steals,	livelihood,	american,	overhyped\\ 
    \rowcolor{black!5}Big Tech & big,	tech,	work,	american,	create,	creators,	everyday,	companies,	systems, americans \\ 
    Artists & artists,	art,	ai,	work,	artist,	human,	people,	creative,	use,	make \\ 
    \rowcolor{black!5}Copyright & copyright,	ai,	copyrighted,	companies,	work,	use,	law,	works,	people,	material\\ 
    Security & \begin{tabular}[c]{@{}l@{}}ai,	data,	security,	research,	systems,	development,	national,	innovation,	policy, government\end{tabular}\\
    \rowcolor{black!5}People's wishes & ai,	people,	don,	want,	resources,	money,	energy,	waste,	jobs,	tool\\ 
    Training data & data,	ai,	scraping,	training,	use,	copyright,	copyrighted,	companies,	scrape,	train \\ \bottomrule
    \end{tabular}
    \caption{Ten top topics in the individuals (IND) subcorpus.}
    \label{tab:topics-individuals}
    \end{small}
\end{table*}

\begin{table*}[t]
\begin{small}
    \centering
    \begin{tabular}{{@{}ll@{}}} \toprule
     \textbf{Interpretation} & \textbf{Words} \\ \midrule
    \rowcolor{black!5}Healthcare & health,	ai,	care,	healthcare,	patient,	clinical,	data,	use,	patients,	medical\\ 
    Tribal-federal & ai,	tribal,	data,	federal,	development,	national,	government,	innovation,	technology,	policy\\ 
    \rowcolor{black!5}\begin{tabular}[c]{@{}l@{}}Research \&  \\ education\end{tabular} & ai,	research,	education,	students,	science,	university,	development,	learning,	data, scientific	\\  
    Policy & \begin{tabular}[c]{@{}l@{}}ai,	human,	american,	policy,	economic,	national,	technology,	development, intelligence,\\	innovation \end{tabular}\\ 
    \rowcolor{black!5} Copyright & copyright,	works,	training,	fair,	use,	ai,	law,	rights,	copyrighted,	creators\\ 
    Energy & energy,	grid,	power,	data,	centers,	center,	demand,	generation,	transmission,	electric \\ 
    \rowcolor{black!5}Security & \begin{tabular}[c]{@{}l@{}}ai,	security,	systems,	national,	government,	federal,	development,	innovation,	american,\\	research \end{tabular}\\ 
    Cybersecurity & ai,	security,	cybersecurity,	systems,	threats,	cyber,	models,	attacks,	threat,	forensic\\ 
    \rowcolor{black!5}Finance & financial,	services,	ai,	banks,	risk,	data,	industry,	management,	regulatory,	use\\ 
    Manufacturing & \begin{tabular}[c]{@{}l@{}}manufacturing,	ai,	energy,	semiconductor,	manufacturers,	industrial,	data,	power,	infrastructure,\\	national \end{tabular}\\ \bottomrule
    \end{tabular}
    \caption{Ten top topics in the REST subcorpus (subcorpora academia, industrial, professional, scientific associations, non-federal organisations, non-government organisations and the private sector).}
    \label{tab:topics-rest}
    \end{small}
\end{table*}

\subsection{Topic modelling}
Topic modelling was performed using BERTopic \cite{grootendorst2022bertopic}, a method relying on pre-trained transformer models and a class-based variation of TF-IDF. The embedding model used was all-MiniLM-L6-v2, topic representations were created with KeyBERTInspired, and the vectoriser was set to CountVectorizer with removal of English stopwords after training. Since the individual responses represent more than 50\% of the corpus, the results were skewed towards over-representing topics in this subcorpus. On the other hand, not all stakeholders contained sufficient documents to conduct topic modelling per stakeholder. For these reasons, topic modelling was performed separately for individuals (IND) and collectively for all other stakeholders (REST). Modelling the REST subcorpora resulted in 15 topics. We additionally performed class-based topic modelling to visualise differences in the frequency of these topics among the rest of the stakeholders. For the individuals, the initial topic modelling resulted in 102 topics, which made interpretation difficult. To focus on the most important topics, we used topic reduction methods, first by setting a minimum cluster size for HDBSCAN \cite{campello2013density} to 25. This resulted in 44 topics, after which we further applied topic reduction without retraining the model, which created a final total of 24 topics for IND.

\subsection{Frequency analysis}
Next, we compared the topics discussed by the six stakeholder groups with those in the AI Action Plan. Because of its size (10,825 tokens), performing topic modelling on this subcorpus is not feasible. For this reason, we performed a frequency analysis using Sketch Engine~\cite{kilgarriff2014sketch}. To keep settings similar to the topic modelling, we used words instead of lemmas and filtered stopwords using the scikit-learn word list. 

To answer RQ2, namely, to what extent the topics discussed by each of the stakeholder groups are reflected in the AI action plan, we combined the frequency analysis with the word representation of the topics. For each of the top 10 topics in the IND and REST subcorpora, we calculated how frequently their most representative words appear in the AI Action Plan. We then summed the word frequencies per topic for the two subcorpora and ranked the topics in descending order. The frequencies show which topics by which stakeholder group are most often mentioned in the AI Action Plan.

\section{Results}

\subsection{Topic modelling}

The top ten topics and the most common words per topic for the individuals (IND) are shown in Table~\ref{tab:topics-individuals}. The individuals mostly discuss topics related to their livelihood and day-to-day life, such as \textit{energy}, the \textit{future} and \textit{security} and commonly refer to AI companies, such as \textit{OpenAI} and in general \textit{Big Tech}.

\begin{figure*}[!t]
    \centering 
    \includegraphics[scale=0.3]{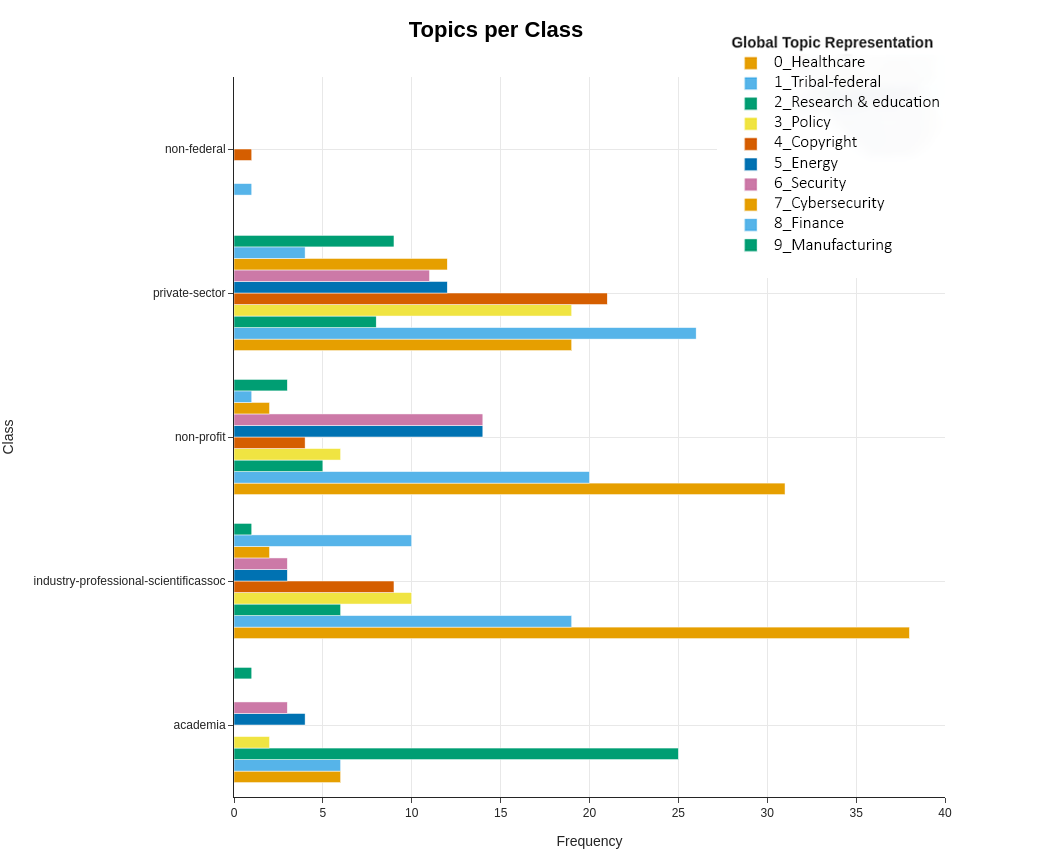}
    \caption{Topics per class for the REST of the respondent types.}
    \label{fig:per-class}
\end{figure*}

From topics and representative words alone, it is clear that individuals hold generally negative opinions of AI and are apprehensive regarding its wider implementation. The second most frequent topic, \textit{stopping theft}, shows a consensus view of AI training as dependent on theft, while 'steal' and 'theft' are recurrent words in several topics. \textit{Copyright} further suggests a negative view of AI in relation to creative works and artistic rights. Other top topics that highlight concerns to lives and livelihood include \textit{energy}, which touches on the high water consumption (see 'water', 'power'), \textit{future}, which suggests AI is 'overhyped' and favours profits over people (see 'place', 'profits', 'steals' and 'livelihood') and \textit{people's wishes} which include protection against job loss (see 'people', 'jobs'). 


The top ten topics and the most common words per topic for the rest of the stakeholders (REST) are shown in Table~\ref{tab:topics-rest}. The mentioned topics here relate to societal issues, including \textit{healthcare}, \textit{governance}, \textit{research \& education} and \textit{policy}. 

The two datasets do have some key terms in common, namely \textit{energy}, \textit{copyright} and \textit{security}. Viewed in context, however, it is clear that each group approaches the topics from a different angle. For example, while \textit{copyright} is discussed by both IND and REST, IND words tend to centre on human aspects (see 'artists', 'work', 'people' and 'creative') while the perspective of REST involves legal aspects ('law') and fair use of works as training data ('fair', 'use'). For REST, more neutral words are used. \textit{Energy} is discussed in technical terms ('grid', 'power', 'data centers'), while IND refer to energy in terms of resources ('water', 'electricity'). One topic that does show some overlap is  \textit{security}, which at a glance shares a focus on national security ('security', 'development, innovation). \textit{Cybersecurity}, which focuses instead on the threat of attacks and the tools for stopping them ('attacks', 'threat', 'forensic') remains an exclusive concern of REST.

Outside of comparisons with IND, it is important to note that REST is far from homogenous and that each of the stakeholder groups tends to have its own priorities. To highlight these priorities Figure~\ref{fig:per-class} shows the distribution of the top ten topics per stakeholder class. The findings shown in Figure~\ref{fig:per-class} are largely in line with expectations. For example, academia is mainly concerned with \textit{Research \& Education}, whereas industry-professional-scientific, a group mostly made up of medical associations, banking associations and technology-related inerest groups are pre-occupied with  \textit{healthcare}, \textit{policies} and \textit{finance}. Societal issues such as \textit{healthcare} and \textit{energy} are discussed among non-profit organisations, together with more regulatory topics such as \textit{security} and \textit{policies}. The private sector shows a stronger focus on topics related to production industries such as \textit{healthcare} and \textit{manufacturing}. \textit{Policy} (both tribal-federal and national) is also frequently discussed, likely reflecting its relevance to the operation and success of private enterprises. \textit{Security} and \textit{cybersecurity} appear prominently as well. Notably, \textit{copyright} emerges as a dominant theme within the private sector, highlighting the critical role of data access and ownership in the development and training of AI systems. 

Overall, while IND prioritise topics related to human life (energy, employment, future, copyright and protection), the interests of the stakeholders in REST are tied to societal, but mostly operational, regulatory, and technological concerns. The prominence of topics like policy, cybersecurity, and copyright underscores a focus on enabling innovation while managing risk and compliance in AI development.

\subsection{Frequency analysis}

\begin{table*}[t]
\begin{small}
    \centering
    \begin{tabularx}{\textwidth}{X}\toprule
    \textbf{Word (frequency)} \\ \midrule
ai (290),
federal (60),
national (52),
doc (49),
led (46),
policy (44),
new (44),
american (43),
systems (42),
security (40),
action (40),
data (39),
plan (38),
infrastructure (38),
america's (38),
actions (35),
development (35),
recommended (30),
ensure (30),
use (30),
government (30),
u.s. (29),
agencies (29),
technology (27),
act (27),
states (27),
models (27),
research (26),
adoption (25),
united (25),
innovation (22),
including (22),
workforce (22),
build (22),
manufacturing (21),
executive (21),
power (21),
intelligence (19),
critical (18),
grid (18),
technologies (18),
dod (18),
energy (17),
export (17),
standards (17),
caisi (17),
science (17),
semiconductor (17),
order (17),
global (16),
existing (16),
tools (16),
office (16),
america (16),
collaboration (15), administration (15), frontier (15), controls (15), relevant (15), information (15), scientific (15), future (15),
department (14), nist (14), programs (14), trump (14), doe (14), u.s.c. (14), create (14),
workers (13), program (13),  advanced (13), defense (13), allies (13), nsf (13), international (13), training (13), develop (13), work (13),
industry (12), efforts (12), response (12), capabilities (12), artificial (12), president (12), centers (12), evaluations (12),
control (11), occupations (11), council (11), education (11), compute (11), access (11), risks (11), establish (11), expand (11), strategic (11), economic (11), promote (10), regulations (10), private (10)
\\ \bottomrule
    \end{tabularx}
    \end{small}
    \caption{Word list of 100 most frequent words in the AI Action Plan.}
    \label{tab:frequency}
\end{table*}

The 100 most frequent words in the AI Action Plan document are shown in Table~\ref{tab:frequency}. They reflect a policy- and strategy-oriented discourse around AI, with a strong emphasis on government, national security, and infrastructure. Words such as 'federal', 'policy', 'government', 'act', 'executive', 'order', 'administration', 'regulations', 'department' point to a high level of regulatory and institutional framing, with attention to the legal and executive context of AI development and deployment. Concerns around national interests and security, geopolitical competition, and AI as a strategic asset are reflected by 'security', 'defense', 'critical', 'controls', 'intelligence', 'export', 'power', 'strategic', 'frontier', 'risks'. A focus on AI-driven economic growth, domestic industrial capacity, and technology leadership is demonstrated by the words 'infrastructure', 'manufacturing', 'semiconductor', 'innovation', 'development', 'industry', 'capabilities', 'centers', 'efforts', 'advanced'. This growth requires a strong emphasis on science, where innovation, R\&D, and human capital development are key pillars of the AI agenda, as shown by the words 'research', 'science', 'technology', 'models', 'training', 'education', 'workforce', 'develop', 'build', 'workers', 'programs'. Lastly, the words 'america’s', 'american', 'united' underscore a national discursive framing, possibly tied to strengthening America's strategic positioning in the AI race.

\begin{figure*}[ht]
    \centering 
    \includegraphics[scale=0.5]{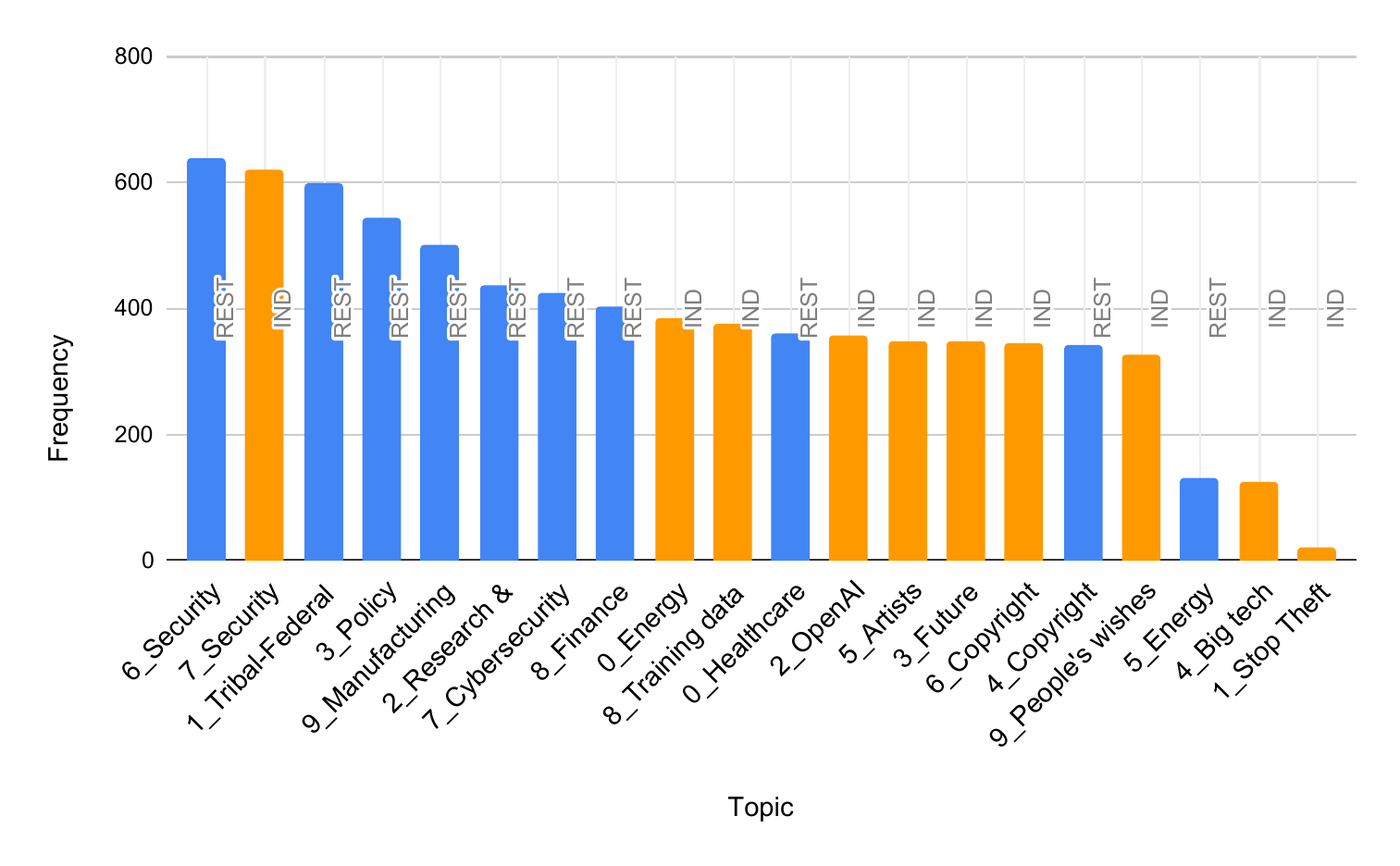}
    \caption{The frequency by which the word representation of the top 10 topics per subcorpora group (IND vs REST) appears in the AI Action Plan. Blue bars represent topics in REST and orange bars topics in IND.}
    \label{fig:freq-per-topic}
\end{figure*}

Figure~\ref{fig:freq-per-topic} shows the total frequency of the words of each topic appearing in the AI Action Plan. The top 10 topics in both REST (blue) and IND (orange) subcorpora are ordered in terms of frequency from highest to lowest. As seen by the blue bars, the topics with the highest word overlap with the AI Action Plan come from the REST subcorpus. The highest overlapping topics are \textit{security, tribal-federal, policy, manufacturing, research \& education, cybersecurity} and \textit{finance}. These topics reflect to a large extent the priorities of the private sector, non-profit organisations and associations. 
In contrast, more societal topics such as \textit{healthcare}, \textit{copyright} and \textit{energy} rank 11th, 16th and 18th. From the IND topics, only \textit{security} shows a high word overlap, ranking second, with all other topics ranking towards the end of the list. This shows that the AI Action Plan has incorporated topics mainly of concern to associations and organisations of the private sector, and much less the concerns expressed by the wider public as expressed in IND.

Cross-examining word frequencies and collocations of specific words between the topics and the AI Action Plan gives some additional observations. \textit{Healthcare}, the largest topic for REST, only appears later in the ranking, with most overlapping words being 'data' (39) and 'use' (30), which are not specific to healthcare. Interestingly, 'healthcare' appears only 3 times in the AI Action Plan, described as a 'critical sector' or an 'economic vertical touched by AI'. The low ranking of the topic of \textit{copyright} and the overlap with the words 'companies' (4), 'work' (13), 'use' (30) and 'training' (13), suggests that the government's priorities, as represented in the AI Action Plan, place relatively little emphasis on copyright-related issues, despite their salience among stakeholders. 
This suggests a gap between the ways the US government frames these issues and stakeholder concerns, particularly in relation to intellectual property and AI training.
Specifically, the word 'copyright' does not appear at all in the AI Action Plan, and 'intellectual
property' only appears once. This may reflect the assumption that copyright is either adequately addressed by existing legal frameworks or is not seen as a major barrier to AI innovation from a policy perspective. 


\section{Discussion}

\textbf{RQ1: What are the prominent topics discussed by each of the six stakeholder groups?}


The topics identified through the analysis reflect the distinct priorities of each stakeholder group. Healthcare and tribal-federal policy dominate responses from non-profits and  associations, aligning with their public service and advocacy roles, while the private sector engages with topics related to innovation and sectoral development, such as policy, copyright, energy, security and manufacturing. Research and education emerge as core concerns for academia, as expected, but are also present in private sector discourse, reflecting shared interests in talent development and collaboration. More than any other stakeholder, individuals are preoccupied with the costs of AI as it relates to energy, loss of jobs, the impact of AI on the future and the violation of the rights of artists to refuse data scraping. These patterns reveal both divergence and overlap in how different actors frame their engagement with AI, shaped by their institutional roles and strategic objectives.

Taken together, these findings illustrate how different stakeholder communities engage with AI from distinct points which are shaped by their missions, capacities, and vulnerabilities. While there is some convergence around topics like security, policy, and innovation, the divergences are equally telling. They highlight the importance of ensuring that AI governance frameworks are not only strategic and forward-looking but also responsive to diverse experiences and sector-specific challenges.


\textbf{RQ2: To what extent are the suggestions and concerns put forth by each of the six stakeholder groups reflected in the AI Action Plan?}

While our work here remains limited to topics alone, the findings do show a clear trend in the overlap between the two datasets and the final AI Action Plan. Though the most referenced topic in the AI Action Plan is \textit{security}, a topic that is shared between IND and REST, the distribution of the remaining 18 topics overwhelmingly favours the ones found in REST. 
Of the ten topics most aligned with the AI Action Plan shown in Figure~\ref{fig:per-class}, 7 are aligned with REST while just 3 are aligned with IND. Of these 7 topics, 'policy', 'research' and 'tribal-federal' fall within the topics most frequently discussed in REST while only a single topic, 'energy', is in the discussed topics from IND. 

The clearest divergence between a topic's presence in the AI Action Plan and its importance to the corpus is \textit{stop theft}, which despite being the second-most-frequently discussed topic in IND shows the least overlap of any of the topics in the AI Action Plan. \textit{Stop theft} and \textit{People's Wishes} are the only topics to mention jobs and general worries around lacking job security and AI redundancies. The position of these topics within the bottom five topics in the AI action plan suggests that these worries remain largely unaddressed by the AI Action Plan.

These findings also reveal potential blind spots in current U.S. AI policy. For instance, copyright (a topic with clear relevance to individuals and civil society) ranks among the lowest in the AI Action Plan. This indicates a misalignment between federal priorities and the real-world concerns of stakeholders. Overlapping words like use, training, law, and rights suggest that copyright is not merely a legal formality but a substantive issue tied to fairness, intellectual ownership, and the ethical development of AI systems. The divergence in emphasis points to the need for more inclusive policy dialogues that account for the interests of less institutionalized voices.

\section{Conclusion}
This paper explores of how different stakeholder groups perceive the social, political, and economic role of AI, drawing on over 10,000 letters submitted in response to the OSTP’s public request. Through the development of a reproducible processing pipeline, we compiled a clean corpus and categorized responses into six stakeholder groups, alongside the 23-page AI Action Plan. Using topic modeling, we identified the dominant themes for each group and compared them with the language and priorities expressed in the AI Action Plan. The findings reveal clear divergences in how stakeholder groups prioritize AI-related issues, with the private sector promoting infrastructure and innovation, non-profits and associations focusing on healthcare and policy, academia emphasizing research and education, and individuals expressing concerns about ethical and societal impacts. While some topics, such as security and innovation, appear across multiple groups, others (like copyright) are under-represented in federal policy despite their prominence in public submissions.

By addressing which topics are emphasized by each group and how these align or fail to align with federal policy, this study highlights the critical role of stakeholder perspectives in shaping national AI agendas. Understanding these dynamics is essential, as public input can influence regulatory decisions with far-reaching consequences. Moreover, the findings emphasize the importance of inclusivity in policymaking, raising important questions about whose concerns are incorporated into national strategies and whose voices are left unheard.

\section{Limitations}
This study is based on a corpus of public submissions to the OSTP, which, while extensive, may not fully represent the broader spectrum of societal perspectives on AI, particularly those of marginalized or under-represented communities who may have lacked access or awareness of the consultation process. Additionally, while topic modelling provides valuable insight into dominant theme, these are influenced by model parameters, preprocessing choices, and the number of topics selected, which may affect the granularity and coherence of the resulting themes. Additionally, the frequency analysis highlights the most commonly used terms but does not account for linguistic patterns, phrase structures, or collocations. We leave the analysis of keywords and collocation patterns for future work.


\section{Bibliographical References}\label{sec:reference}

\bibliographystyle{lrec2026-natbib}
\bibliography{lrec2026-example}


\end{document}